\documentclass[letterpaper]{article} 
\usepackage{aaai25}  
\usepackage{times}  
\usepackage{helvet}  
\usepackage{courier}  
\usepackage[hyphens]{url}  
\usepackage{graphicx} 
\usepackage{natbib}  
\usepackage{caption} 
\usepackage{algorithm}
\usepackage{algorithmic}

\usepackage{newfloat}
\usepackage{listings}
\DeclareCaptionStyle{ruled}{labelfont=normalfont,labelsep=colon,strut=off} 
\floatstyle{ruled}
\newfloat{listing}{tb}{lst}{}
\floatname{listing}{Listing}
\usepackage{booktabs}
\usepackage{comment}
\usepackage{enumitem}
\usepackage{colortbl}
\usepackage{pdflscape}
\usepackage{xcolor}
\usepackage{array}

\definecolor{codegreen}{rgb}{0,0.6,0}
\definecolor{codegray}{rgb}{0.5,0.5,0.5}
\definecolor{codepurple}{rgb}{0.58,0,0.82}
\definecolor{backcolour}{rgb}{0.95,0.95,0.92}
\definecolor{codelavender}{RGB}{230, 230, 250}
\definecolor{intgray}{RGB}{233, 239, 240}
\definecolor{purple}{RGB}{129,1,126}
\definecolor{infra-white}{RGB}{134,206,234}

\newcommand{\name}{\textsc{WorldAPIs}}
\newcommand{\grayout}[1]{{\textcolor{gray}{#1}}}
\newcommand{\mytexttt}[1]{{\colorbox{codelavender}{\mystrut(.5, .5)\texttt{#1}}}}
\def\mystrut(#1,#2){\vrule height #1pt depth #2pt width 0pt}

\lstdefinestyle{mystyle}{ 
    commentstyle=\color{codegreen},
    keywordstyle=\color{magenta},
    numberstyle=\tiny\color{codegray},
    stringstyle=\color{codepurple},
    basicstyle=\ttfamily\footnotesize,
    breakatwhitespace=false,         
    breaklines=true,                 
    captionpos=b,                    
    keepspaces=true,                 
    numbers=left,                    
    numbersep=5pt,                  
    showspaces=false,                
    showstringspaces=false,
    tabsize=2
}

\usepackage{array}
\newcommand{\PreserveBackslash}[1]{\let\temp=\\#1\let\\=\temp}
\newcolumntype{C}[1]{>{\PreserveBackslash\centering}p{#1}}
\newcolumntype{R}[1]{>{\PreserveBackslash\raggedleft}p{#1}}
\newcolumntype{L}[1]{>{\PreserveBackslash\raggedright}p{#1}}

\lstdefinestyle{myCustomPythonStyle}{
  language=python,
  numbers=left,
  basicstyle=\scriptsize\ttfamily,
  stepnumber=1,
  showspaces=false,
  showstringspaces=false
}
\title{\name{}: 
The World Is Worth How Many APIs? \\ A Thought Experiment}
\author{
    Jiefu Ou,~
    Arda Uzunoğlu,~
    Benjamin Van Durme,~
    Daniel Khashabi~
}
\affiliations{
    Johns Hopkins University\\

    \{jou6, auzunog1, vandurme, danielk\}@jhu.edu
}

\begin{document}

\maketitle

\begin{abstract}
AI systems make decisions in physical environments through primitive actions or affordances that are accessed via API calls. 
While deploying AI agents in the real world involves \emph{numerous} high-level actions, existing embodied simulators offer \emph{a limited set} of domain-salient APIs. 
This naturally brings up the questions: \emph{how many primitive actions (APIs) are needed for a versatile embodied agent, and how should they look like?}

We explore this via a thought experiment: assuming that wikiHow tutorials cover a wide variety of human-written tasks, what is the space of APIs needed to cover these instructions? 
We propose a framework to iteratively induce new APIs by grounding wikiHow instruction to situated agent policies. 
Inspired by recent successes in large language models (LLMs) for embodied planning, we propose a few-shot prompting to steer GPT-4 to generate Pythonic programs as agent policies and bootstrap a universe of APIs by 1) reusing a seed set of APIs; and then 2) fabricate new API calls when necessary. 
The focus of this thought experiment is on defining these APIs rather than their excitability.

We apply the proposed pipeline on instructions from wikiHow tutorials. On a small fraction (0.5$\%$) of tutorials, we induce an action space of 300+ APIs necessary for capturing the rich variety of tasks in the physical world.
A detailed automatic and human analysis of the induction output reveals that the proposed pipeline enables effective reuse and creation of APIs. 
Moreover, a manual review revealed that existing simulators support only a small subset of the induced APIs (9 of the top 50 frequent APIs), motivating the development of action-rich embodied environments.
\end{abstract}

%

\section{Introduction}
\begin{figure*}[ht]
\centering
\includegraphics[scale=0.089,trim=0cm 0.5cm 0cm 0cm,width=0.89\textwidth]{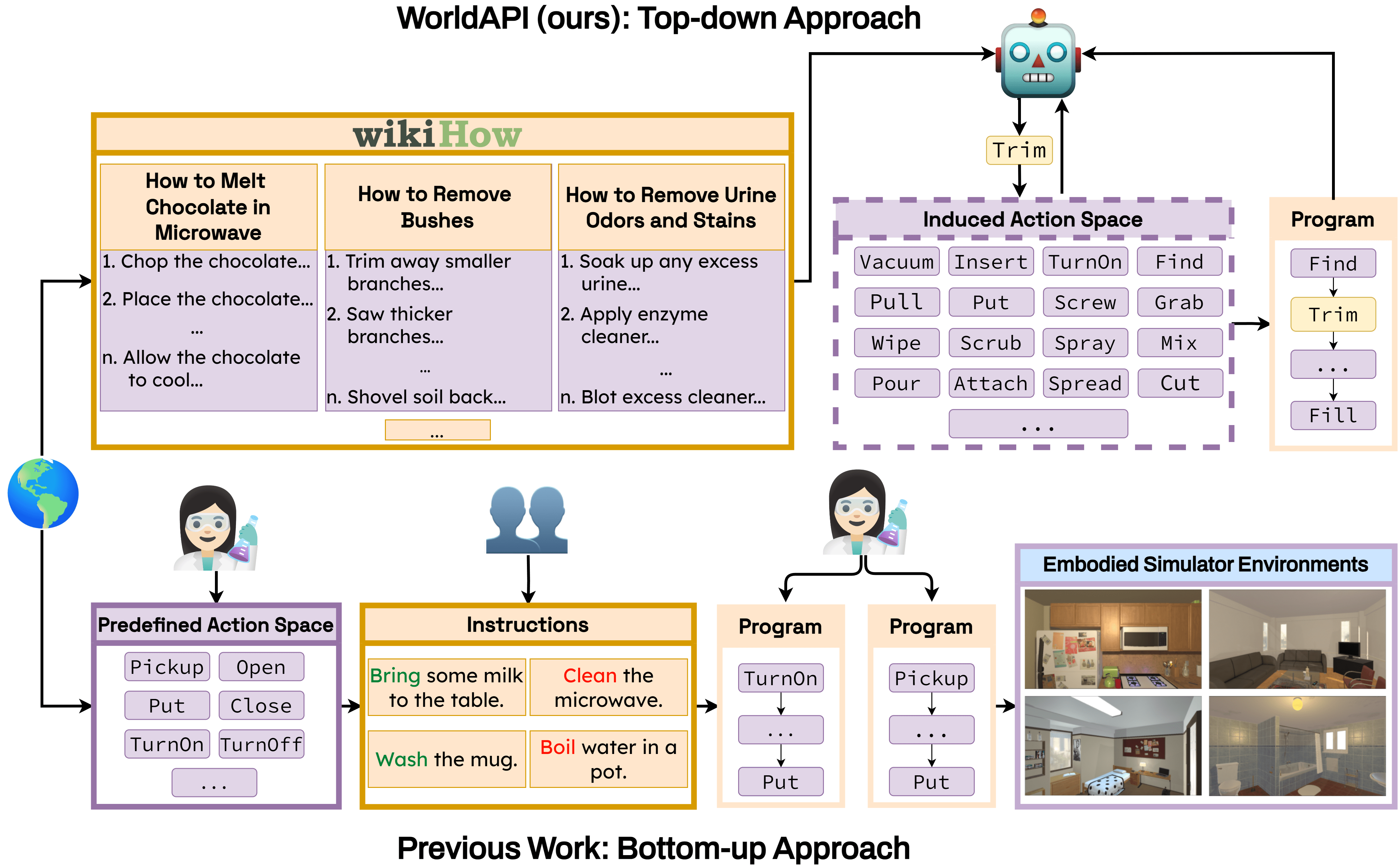}
    \caption{\textbf{Top}: \name, the proposed thought experiment that takes a top-down approach. Starting from daily tasks with sequences of instruction steps in wikiHow, and a seed action space (API pool), we iteratively prompt LLMs to generate agent programs and add the induced (hallucinated) APIs in generated programs to the API pool.
    \textbf{Bottom}: in contrast, most of the prior work in building embodied environments often adopts a bottom-up approach. The simulation and collection of instructions and programs are all based on a close set of predefined actions.}
    \label{fig:intro}
\end{figure*}

Developing versatile and capable AI agents that follow natural language instructions for task execution in the real world has been an important research objective. A series of efforts \citep{Ahn2022DoAI, Singh2022ProgPromptGS, song2023llmplanner} have been made towards this goal, with a notable upsurge of work on harnessing the power of large language models (LLMs) to generate task plans and policies for embodied agents.

The backbones of such success are benchmarks and simulations of embodied environments.  \citep{anderson2018visionandlanguage, thomason2019shifting, Puig2018VirtualHomeSH, alfred, ku2020roomacrossroom}. However, most of these existing embodied environments suffer from the limited hand-crafted action spaces. Such restraint diminishes action diversity, leading to the oversaturation of a small group of basic actions. For example, as shown in Figure \ref{fig:intro}, existing embodied simulators/environments often define a small and closed set of actions an agent can perform.
As a result, the possible instructions it can carry out are highly constrained by the limited action space.  
However, carrying out tasks in real-world scenarios across different domains requires a much larger and more diverse action space.
To bridge this gap between embodied agent learning and real-world deployment, it is natural to ask the following questions: \emph{To build a versatile embodied agent that can carry out daily tasks in the physical world, how many primitive actions (APIs) should such an agent be equipped with, and what do they look like?}

In this work, we make the first step towards answering these questions by proposing \name (\S\ref{sec:worldapi}), a thought experiment that aims to approximate a realistic pool of agent-executable primitive actions (APIs)\footnote{In this work we refer to APIs as interfaces to affordances of agents that could be deployed in the real world. Upon calling an API, the agent can carry out a corresponding physical primitive action. We use APIs and primitive actions and APIs interchangeably henceforth}, the composition of which enables agents to accomplish diverse and realistic goals specified by high-level procedural natural language instructions. The thought experiment assumes an embodied agent to be deployed in a hypothetical world consisting of objects and primitive action space with predefined properties. The agent is tasked to perform a wide variety of daily tasks specified through natural language instructions, in the form of wikiHow tutorials. To complete these tasks, the agent needs to evoke and compose primitive actions (APIs) from the action space to interact with objects. As shown in Figure \ref{fig:intro}, in contrast to prior work, we propose to induce an open action space from diverse and realistic wikiHow procedural instructions. 

To create the action space, we propose a pipeline (\S\ref{sec:pipeline}) for bootstrapping the API pool and corresponding agent policies as Pythonic programs that 1) iteratively generate semi-executable agent policies as Pythonic programs via few-shot prompting LLMs with procedural natural language tutorials. 2) parse generated programs, filter and add full/snippets of programs to the pool of demonstrations. The proposed pipeline steers the LLMs to \textit{1) reuse existing APIs when possible} and \textit{2) hallucinate (induce) new APIs when the instruction cannot be covered with existing APIs} when converting a natural language instruction into a composition of APIs. 
While hallucination is often regarded as a pitfall of LLMs, in this work we harness it as a constructive capability to synthesize novel APIs that are semantically and stylistically consistent with our pre-defined APIs.

We conduct preliminary experiments (\S\ref{sec:exp}) to induce agent policies and action spaces on wikiHow \footnote{\url{https://www.wikihow.com}}, a rich semi-structured online resource of human-written procedural instructions covering daily tasks from a wide spectrum of domains. The set of APIs saturated after running the iterative induction pipeline on less than $0.5\%$ of wikiHow tutorials, yielding a self-contained pool of over 300 frequently-evoked APIs (\S\ref{subsec:auto_eval}). This provides an approximation of the lower bound of the primitive action space. Human and automatic evaluations (\S\ref{subsec:human_eval}) demonstrate the proposed pipeline's effectiveness in inducing APIs necessary to carry out diverse wikiHow instructions. Further comparison with existing embodied environments \citep{alfred, Puig2018VirtualHomeSH} highlights the limited coverage of our induced APIs in existing environments (\S\ref{subsec:auto_eval}), thus encouraging further efforts to enrich the action space of embodied simulations.

\section{Related Work}
\paragraph{Embodied planning and simulation.} 
This literature focuses on enabling agents to navigate and manipulate their environments. While LLMs are shown to be effective at planning~\citep{song2023llmplanner, huang2022language, Huang2023GroundedDG}, embodied planning usually requires more than textual data due to its interactive nature. Thus, creating a reflection of the real world (e.g., simulation or benchmarks) has been an important aspect of the progress. 
These simulators enable interactive learning and evaluation on a proxy of the world.  
 
Various simulators have been developed for simulating human activities \citep{Kolve2017AI2THORAI, xia2018gibson, Nasiriany2024RoboCasaLS}, each with distinct advantages. Yet, they are intrinsically confined to a set of interactions based on factors such as the feasibility of rendering actions, computational limitations, constraints of the physics engine, etc. 
These benchmarks for embodied reasoning (summarized in Table \ref{tab:datasets}) can be split into two groups: (i) grounded in \emph{physical} world and (ii) grounded in \emph{software} world. 

\begin{table}
\resizebox{\columnwidth}{!}{%
\begin{tabular}{L{5.4cm}C{2cm}R{3.7cm}}
        \toprule
        Resource & Size of Action Space & Domain \\
        \midrule
        \cellcolor{intgray} \textsc{Physical World Simulators} & \cellcolor{intgray} & \cellcolor{intgray} \\
        ALFRED \citep{alfred} & 13 & Household Tasks \\
        VirtualHome \citep{Puig2018VirtualHomeSH} & 12 & Household Tasks \\
        TEACh \citep{padmakumar2021teach} & 16 & Household Tasks \\
        LACMA \citep{Yang2023LACMALC} & 10 & Household Tasks \\
        BEHAVIOR \citep{Srivastava2021BEHAVIORBF} & 6 & Household Tasks \\
        BEHAVIOR-1K \citep{Li2024BEHAVIOR1KAH} & 6 & Household Tasks \\
        Mini-BEHAVIOR \citep{Jin2023MiniBEHAVIORAP} & 15 & Household Tasks \\
        House3D \citep{Wu2018BuildingGA} & 12 & Navigation \\
        R2R \citep{anderson2018visionandlanguage} & 6 & Navigation \\
        CVDN \citep{thomason2019visionanddialog} & 6 & Navigation \\
        HM3D \citep{Ramakrishnan2021HabitatMatterport3D} & 4 & Navigation \\
        ThreeDWorld \citep{Gan2021TheTT} & 6 & Navigation,Manipulation \\
        CHAI \citep{misra2019mapping} & 5 & Navigation,Manipulation \\
        CALVIN \citep{Mees2021CALVINAB} & 12 & Manipulation \\
        Overcooked \citep{Carroll2019OnTU} & 6 & Cooking \\
        VRKitchen \citep{Gao2019VRKitchenAI} & 6 & Cooking \\
        KITchen \citep{Younes2024KITchenAR} & 12 & Household \& Cooking \\
        SmartPlay \citep{Wu2023SmartPlayA} & 37 & Games \\
        \midrule
        \cellcolor{intgray} \textsc{Digital Simulators} & \cellcolor{intgray} & \cellcolor{intgray}  \\
        \grayout{ToolAlpaca \citep{Tang2023ToolAlpacaGT}} & \grayout{400} & \grayout{Function Call} \\
        \grayout{API-Bank~\citep{li2023apibank}} & \grayout{53} & \grayout{Function Call} \\
        \grayout{API-Bench \citep{Patil2023GorillaLL}} & \grayout{1.6K} & \grayout{Function Call} \\ 
        \grayout{API-Pack \citep{Guo2024APIPA}} & \grayout{11.2K} & \grayout{Function Call} \\
        \grayout{ToolBench~\citep{Xu2023OnTT}} & \grayout{232} & \grayout{Function Call} \\
        \grayout{ToolLLM \citep{Qin2023ToolLLMFL}} & \grayout{16.4K} & \grayout{Function Call} \\
        \midrule
        \textbf{\name{} (ours)} &  \textbf{300+} & \textbf{Household Tasks \& Cooking \& Crafts, etc} \\
        \bottomrule
    \end{tabular}%
}
\caption{Existing resources, their sizes and domains.}
\label{tab:datasets}
\end{table}

The first group of simulators is concerned with various real-world tasks such as navigation. In most of these works, plan execution resources are mostly limited to household and adjacent domains and rely on a hand-crafted small action space.   Such limitations of the action space diminish the diversity of the tasks that can be executed with the given actions. 
Thus, they fail to capture the authenticity of real-world tasks due to their dependency on an extremely restrictive set of actions and incomplete modeling of the real world in virtual environments. 
In contrast, here, we ignore the concerns regarding the action (API) executability to allow us to broadly explore the space of various actions needed in the world.  

The second category of benchmarks deals with executing APIs in digital spaces (e.g., interacting with web pages) without correspondence in the real world.
We have highlighted notable works in this category, even though the focus of our thought experiment is on APIs needed to interact with the physical world.

\paragraph{Procedural language understanding.} There is a broad range of works that focus on \emph{procedures}, including reasoning on procedures ~\citep{zhang-etal-2020-reasoning, uzunoglu2024paradise}, 
generating procedures~\citep{sakaguchi-etal-2021-proscript-partially, lyu-etal-2021-goal}, and so forth. 
Most of this literature is confined to unimodal tasks, and the multimodal approaches are limited \citep{zhou2023procedureaware, zellersluhessel2021merlot}.
Similarly, the focus has been on individual tasks, even though collective benchmarks that cover different tasks exist \citep{uzunoglu-sahin-2023-benchmarking, onoe2021creak} 
Therefore, most works are confined to shallow textual and visual representations. This leads to the lack of procedural data grounded in real-world tasks and scenarios. Similar to what we aspire to achieve, \citet{Puig2018VirtualHomeSH} utilizes wikiHow to convert high-level procedures to executable actions in virtual environments. However, its breadth is highly narrow due to its limited domain and predefined action space of 12 actions.

\section{Defining a Hypothetical World}
\label{sec:worldapi}
Our goal is to formulate simulations that allow us to approximate the action space of versatile robots physical world.

Unlike prior work that takes a \textit{bottom-up} approach (i.e., first define the action space and then build the simulator), we adopt a \textit{top-down} formalism: we first collect diverse and realistic instructions from online resources (\S\ref{subsec:wikihow}), then define \textit{hypothetical} environment and agent that are capable of carrying out these instructions (\S\ref{subsec:envrionment}), and finally induce agent programs and action spaces jointly (\S\ref{sec: method}). 

\subsection{Collecting instructions from wikiHow}
\label{subsec:wikihow}
We leverage wikiHow, a prominent web platform with 200K+ professionally curated ``how-to" tutorials across a diverse set of domains.
Each wikiHow tutorial presents a sequence of instructional paragraphs in natural language that aim to accomplish a certain goal (see Figure \ref{fig:intro} for an example). We follow prior work \citep{zhang-etal-2020-reasoning, zhou-etal-2022-show} to use the tutorial title as the goal, the paragraph headline as instruction steps, and the paragraph body as additional descriptions.

Crucially, wikiHow is curated to be used \emph{by humans}. It contains a significant diversity of tasks that require seemingly similar actions (cutting vs. pruning, tying vs. fastening, etc.) with subtly different physical control specifications,
making it a rich source of tasks for simulating embodied agents of the real world.

While wikiHow covers a wide range of domains, we choose \emph{Home \& Garden} as our domain due to its feasibility to be a test bed for embodied agents and its ample action space reflective of the real world. Thus, we exclude categories that are abstract and social (i.e. \textit{Relationships} and \textit{Youth}), contain actions that would not be expected to be performed by embodied agents (i.e. \emph{Work World}), or have repetitive actions that fail to illustrate the scale of realistic action space (i.e. \emph{Computers \& Electronics}).

\subsection{A Text-based Hypothetical Environment based on wikiHow Instructions}
\label{subsec:envrionment}
In our formulation, a text-based hypothetical environment is defined by each wikiHow tutorial.
The instructions in the tutorials provide all the objects explicitly
or implicitly required to complete the task inferred by commonsense knowledge.

Since we are not building a real simulator, we make the following simplifications to the environment: 1) Scenery and spatial details are omitted and simplified to object receptacle relations. For example, liquid/powder is within the container; 2) As a result of spatial simplification, navigation, and object detection are simplified to a \mytexttt{find} API, which directs the agent to locate and navigate to the vicinity (within intractable distance) of a target object.

\subsection{Principles for Translating Language Instructions to Pythonic Policies and Actions}
Inspired by recent work~\citep{Liang2022CodeAP, Singh2022ProgPromptGS}, we jointly induce primitive actions (APIs) and policies (Pythonic programs) via prompting LLMs with few-shot demonstrations (details in \S\ref{sec: method}). The demonstrations provide information about the hypothetical environment to the LLMs: available objects, primitive actions (APIs), as well as how to interact with objects through API calls and state checking. We create an annotation guideline that defines the semantic formalism of objects and APIs for the hypothetical environment. We annotate the programs for a small set of wikiHow tutorials as seed demonstrations.

\paragraph{Overview.}
The example in Listing \ref{lst:annotation} shows a demonstration that translates the first wikiHow tutorial in the top-left of Figure \ref{fig:intro}, i.e. ``How to Melt Chocolate in Microwave''. After specifying the tutorial (\mytexttt{TASK}) and numbered instruction steps (\mytexttt{INSTRUCTIONS}), the following \mytexttt{PROGRAM} lists the objects and primitive actions through \mytexttt{import}, specifies object relations with \mytexttt{receptacles}, and incorporates instruction steps and sub-steps in comments for policy specification.

\begin{listing}[tb]%
\caption{Our in-context demonstrations for decomposing wikiHow tasks into API calls.}%
\label{lst:annotation}
\begin{lstlisting}[language=Python, style=myCustomPythonStyle]
TASK: 
How to Melt Chocolate in Microwave
INSTRUCTIONS:
1. Chop the chocolate ...
2. Place the chocolate ...
...
PROGRAM:
# primitive APIs
from utils import find, grab, put, put_back ...
# objects
from objects_pool import chocolate_0, knife_0 ...
# object-object relation specification
faucet_0.receptacles.append(sink_0)
...
# program
def robot_program():
    # 1. Chop the chocolate into small pieces with a serrated knife.
    # find and grab chocolate
    find(obj=chocolate_0)
    grab(obj=chocolate_0)
    # find and grab knife
    find(obj=knife_0)
    grab(obj=knife_0)
    # find cutting board
    find(obj=cutting_board_0)
    # chop chocolate until it's into small pieces
    while chocolate_0.material_properties[`form'] != `small pieces':
        chop(obj=chocolate_0, tool=knife_0, on=cutting_board_0)
    put_back(obj=knife_0)
    # 2. Place the chocolate into a microwave-safe bowl.
    ...
# execute the program
robot_program()
\end{lstlisting}
\end{listing}

\paragraph{Defining objects.}
Following \citet{Kolve2017AI2THORAI}, we unify all the objects under the \mytexttt{WorldObject} class that is defined with two types of properties as described below:

\underline{Actionable properties} specify the APIs an agent can invoke to interact with an object. For example, the \mytexttt{grab} API can only target objects with the actionable property \mytexttt{grabbable}.

\underline{Material properties} refer to the inherent attributes of objects that are not directly tied to interactions but can be altered through API calls. For instance, the chocolate object in Listing\ref{lst:annotation} with the material property \mytexttt{\{`form': `bar'\}} may change to \mytexttt{\{`form': `small pieces'\}} after a \mytexttt{chop} API call.

\underline{Object-object relationships} can be specified via \mytexttt{receptacles} and \mytexttt{receptacles\_of}, when necessary. 
We refer readers to apendix on our arXiv draft for the complete definition of \mytexttt{WorldObject}. We refer readers to  \S\ref{subsec_appx: define objects} for the complete definition of \mytexttt{WorldObject}.

\paragraph{Defining primitive actions (APIs).}
While there are numerous potential ways to define the action spaces, we seek to strike a balance in our definition among 1) stylistically and semantically similar to existing embodied environments \citep{alfred}; 2) simple and informative such that LLMs can effectively leverage its parametric knowledge of code generation to predict agent policies as programs for unseen instructions; 3) easily extendable where LLMs are capable of hallucinating new APIs that are consistent to the available APIs specified in the demonstrations. Specifically, we focus on the following aspects:

\underline{Base APIs:} To ensure comparability with prior work, we provide a set of commonly used APIs for executing wikiHow instructions, such as \mytexttt{find}, \mytexttt{put}, \mytexttt{open\_obj}, and \mytexttt{turn\_on}. Details are in \S\ref{subsubsec_appx: base apis}.

\underline{Granularity:} 
Defining the right abstraction level for actions is challenging. Nevertheless, in our definitions, we aim to 
strike a balance between the following two principles: (Further details on these principles in \S\ref{subsubsec_appx: granularity})

\begin{itemize}[leftmargin=10pt]
    \item Avoiding overly-abstract actions: 
    We prefer to have tasks broken into actions that directly interact with the objects.
    For example, given the task of ``dry your clothes with a drier,'' one straightforward approach is distilling it to a single \mytexttt{dry} action. However, we argue that it is not a desired approach to create the action space since it creates APIs at a problematic level of granularity from an action perspective - drying something with a dryer requires a totally different set of actions to drying something with iron. 
    Instead, break up this task into sub-steps (e.g., ``the dryer can be opened and turned on to perform drying''), each appropriate action. 
    \item  Avoiding too low-level actions: 
    We prefer to avoid actions that involve low-level physics. As this degrades the action space into just a handful of control APIs with complicated spatial-temporal argument realization, which is infeasible in our hypothetical environment without explicit spatial relation specification, and harms the reusability of learned APIs in practice. For instance in Listing \ref{lst:annotation}, given the instruction substep ``chop chocolate until it ’s into small pieces'', we will define \mytexttt{chop(obj=chocolate\_0, tool=knife\_0,} \mytexttt{ on=cutting\_board\_0)}, instead of \mytexttt{move\_held\_object(moveMagnitude=0.1)}.
\end{itemize}

\underline{State checking and feedback loop:}
Similar to prior work \citep{Singh2022ProgPromptGS}, 
in our demonstrations we implement steps where the agent collects feedback from the environments by checking the object state (material properties) and acting accordingly (details in \S\ref{subsubsec_appx: feedback})

\underline{Actionability:}
At times we might encounter instructions that are either too broad, subjective, or under-specified to be programmed by any existing/new APIs. For example, actions that involve personal feelings, or subjective preferences. 
Whenever we encounter such instructions, we skip them with a comment \mytexttt{\# skip this instruction}.

\section{Inducing the Action/Policy Space in the Hypothetical World}\label{sec:pipeline}
\label{sec: method}
\begin{figure*}[ht]
    \centering
    \includegraphics[scale=0.86,trim=0cm 9.8cm 5.9cm 1.0cm,clip]{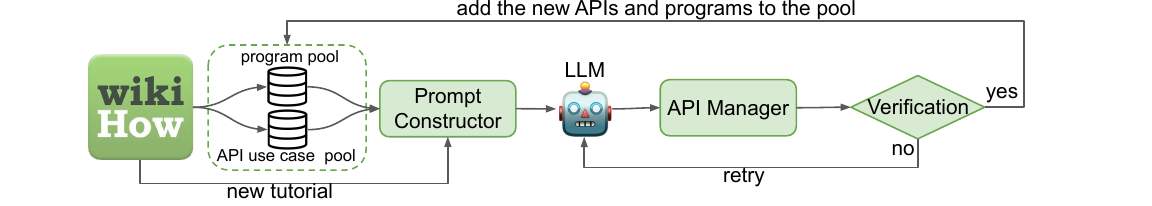}
    \caption{Proposed pipeline that jointly induces new APIs and programs.}
    \label{fig:pipeline}
    \vspace{-0.1cm}
\end{figure*}

Following our definition of the hypothetical world (\S\ref{sec:worldapi}), we develop a pipeline for inducing action/policy of wikiHow tutorials via
iterative few-shot code generation with LLMs. 
As depicted in Figure \ref{fig:pipeline}, at each step of induction, a random tutorial 
is sampled from wikiHow. 
Given the input tutorial, a prompt is constructed with a system instruction, retrieved programs, and API use cases that are used as demonstrations to guide LLM generation. 
The LLMs process this prompt, after which we verify the syntactic well-formedness of the generated program. 
If it passes the verification, we add the full program and the extracted API into the pool of demonstrations. 
This is done iteratively, monotonically expanding the pool of APIs/programs. At each round,  LLM leverages the program examples generated by itself in previous steps, essentially bootstrapping from its prior output.

\paragraph{Prompt construction.}
The prompts provided to LLM start with a system instruction that specifies the task requirements 
and the \mytexttt{WorldObject} definition. The system prompt is followed by a sequence of $k$ (we use $k=10$) in-context examples. These examples are pairs of instructions and their corresponding programs, retrieved based on the semantic similarity to a particular instruction in the given tutorial (see Figure \ref{fig:pipeline:appendix} for an abridged example).\\ 

\vspace{-0.4cm}
\begin{listing}[h]%
\caption{Example API use cases}%
\begin{lstlisting}[language=Python, style=myCustomPythonStyle]
# Use Case of squeeze
# bring the sponge to the sink and squeeze out the water
    find(obj=sink_0)
    while sponge_0.material_properties['filled'] != 'empty':
        squeeze(obj=sponge_0, target=sink_0)

# Use Case of insert
# attach hose attachment to vacuum
insert(obj=hose_attachment_0, target=vacuum_0)
\end{lstlisting}
\end{listing}

\paragraph{API use cases.}
To provide more complete information about all existing APIs, we prepend use cases of all APIs discovered thus far to the input prompt. The use cases are the most similar use cases of each API retrieved from the pool. The sequence of use cases is inserted before the full programs as extra demonstrations. The use case of API $f$  is a program snippet starting from a sub-step comment and ending at the line that calls $f$. 

\paragraph{Expanding the pool of demonstrations and APIs.}
\label{subsec:expansion}
We perform rejection sampling on program generation. Specifically, a program that either contains syntactic errors (i.e. failed in Abstract Syntax Tree parsing) or does not fully follow the instructions (i.e. does not contain comments of every numbered instruction step) will be rejected and the LLM will regenerate the program. For all the verified programs, we extract the full program and all the code snippets as API use cases and add them to the corresponding pools.

\paragraph{Leveraging action descriptions in wikiHow.}
\label{subsubsec:description}
While most of the prior work only uses the headline of each instructional paragraph in wikiHow, we find that their descriptions contain beneficial details 
to decompose and ground the instructions into primitive actions. For example, the header ``\textit{Combine the sugar, cocoa powder, and salt in a saucepan}'' omits the action of mixing the ingredients that are specified in the description body ``\textit{Pour the sugar into a small saucepan. Add the unsweetened cocoa powder and a dash of salt. Stir everything together with a whisk.}''
We, therefore, include these descriptions as part of the input instructions.

\begin{figure}[ht]
    \centering
    \includegraphics[scale=0.26,trim=0.4cm 0.8cm 0cm 0cm, width=0.45\textwidth]{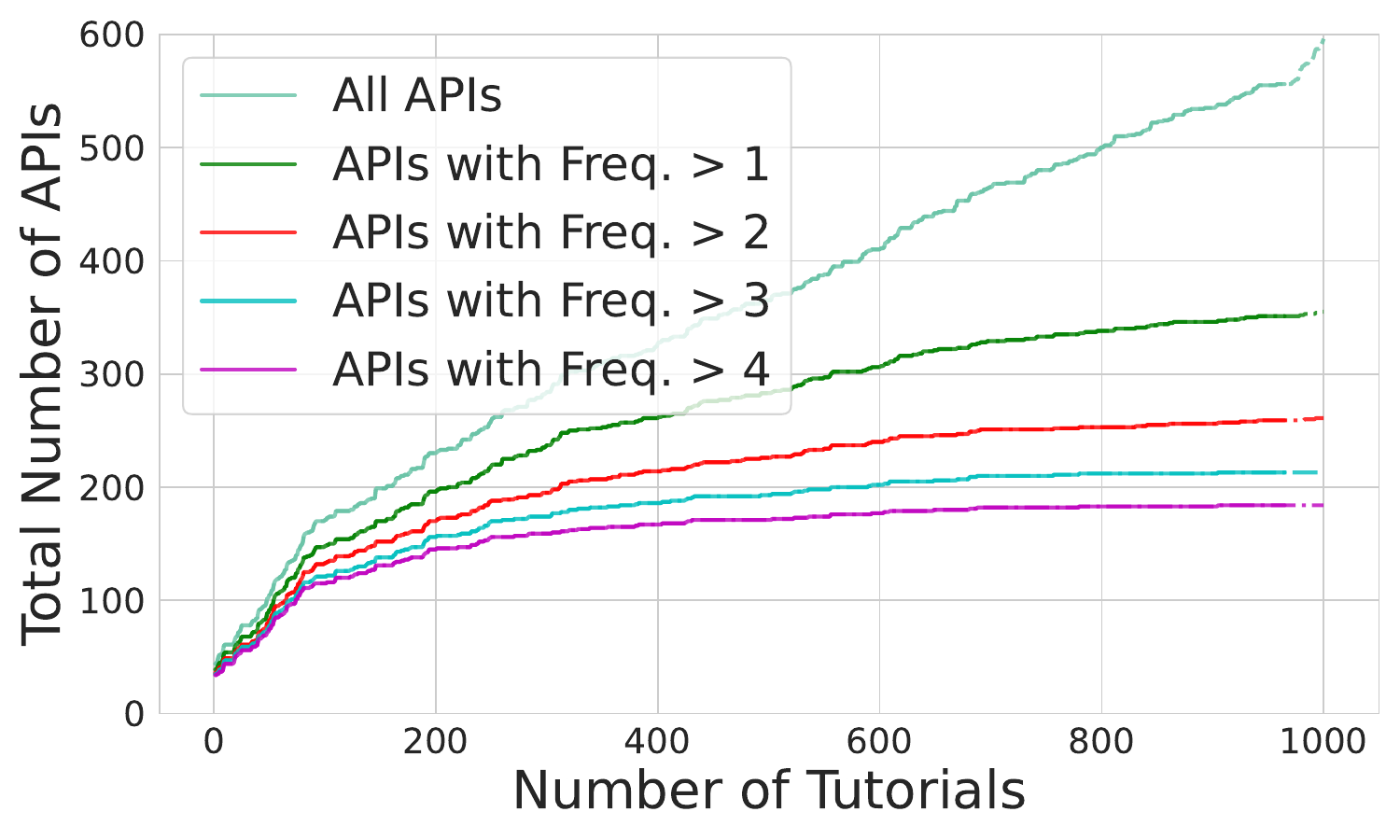}
    \caption{Size of API pool vs. \# of tutorials. Lines represent different frequency thresholds used to filter the APIs. With increasing thresholds, the size of the pool of frequently-evoked APIs stabilized after inducing $\sim$ 600 tutorials.
    }
    \label{fig:freq_based}
\end{figure}

\section{Experiment \& Evaluation Metrics}
\label{sec:exp}
We apply the proposed induction pipeline to the Home and Garden category of wikiHow tutorials.
\subsection{Experimental Setup}
We perform human annotation on 19 randomly sampled tutorials according to the annotation guideline, which serves as the seed demonstrations. We then sample 1000 random tutorials to represent the diverse procedural instructions from users in real-world scenarios. Then, we evaluate 3 variants of our pipeline on the sampled set:
\begin{itemize}[leftmargin=10pt]
    \item The \textbf{Base} variant takes in only pairs of (instruction steps, full program) as in-context demonstrations.
    \item The \textbf{Base + Use Case} version that additionally takes in code snippets of API use cases as demonstrations.
    \item The \textbf{Base + Use Case + Description} version that includes API use cases in demonstrations and adds descriptions to each instruction step (discussed in \S\ref{subsubsec:description}).
\end{itemize}

For all the pipeline variants, we use OpenAI \texttt{gpt-4-1106-preview} as the backbone LLM, with temperature = 0. In prompt construction, we use top $k=10$ similar full-program and use the top-1 similar use case for every API provided/induced so far at each step as in-context demonstrations. When selecting demonstrations, we measure retrieval similarity based on the embedding similarity of the concatenation of the tutorial name and instruction steps of the full program demonstrations and the leading comment of the API use case demonstrations. We use OpenAI \texttt{text-embedding-ada-002-v2} to obtain text embeddings. 

We run all the pipeline variants for the first 50 tutorials within samples for human evaluation and apply \textbf{Base + Use Case + Description} to induce the action space on the full set of 1000 tutorials to induce the action space.

\begin{table*}[ht]
\centering
\small
\begin{tabular}{l|ccc|cc|cc}\toprule
              & \multicolumn{3}{|c|}{Redundancy$\downarrow$}                                                  & \multicolumn{2}{|c|}{Faithfulness$\uparrow$} 
             & \multicolumn{1}{c}{APIs} &  \\
             \midrule
Induction Pipelines      & Score & -Complex & \begin{tabular}[c]{@{}c@{}}-Complex \\ -Synonym\end{tabular} & Score          & Ranking         & Avg. \# \\
\midrule
Full (a)      & 46.50 & 38.11    & 35.32                                                        & 82.0           & 1.756           & 2.88    \\
+UseCase      & \textbf{43.44} & \textbf{36.07}    & 34.43                                                        & 81.0           & 1.732           & 1.24     \\
+UseCase+Desc & 47.46 & 36.59    & \textbf{33.70}                                                        & \textbf{84.0}           & \textbf{1.439}           & 1.74      \\
\bottomrule
\end{tabular}
\caption{Human evaluation results on the output of 50 wikiHow tutorials. For redundancy, ``Score'' is the full score, and  ``-Complex''/``-Synonyms'' refers to rescoring all the new APIs that are too complicated to be further decomposed/synonyms to existing APIs from 0.5 (partially redundant) to 1 (fully use full), respectively. For faithfulness, ``Score'' is the absolute score, and ``Rank'' is the preference-based ranking. ``Avg. \#'' of APIs lists the average number of new APIs induced per tutorial. \textbf{Adding API use case demonstrations helps decrease redundancy and significantly reduce the creation of new APIs, and further adding descriptions improves faithfulness.}}
\label{tab:qual_new}
\end{table*}
\begin{figure*}[ht]
    \includegraphics[scale=0.38,trim=0.25cm 0.8cm 0cm 0cm,width=\textwidth]{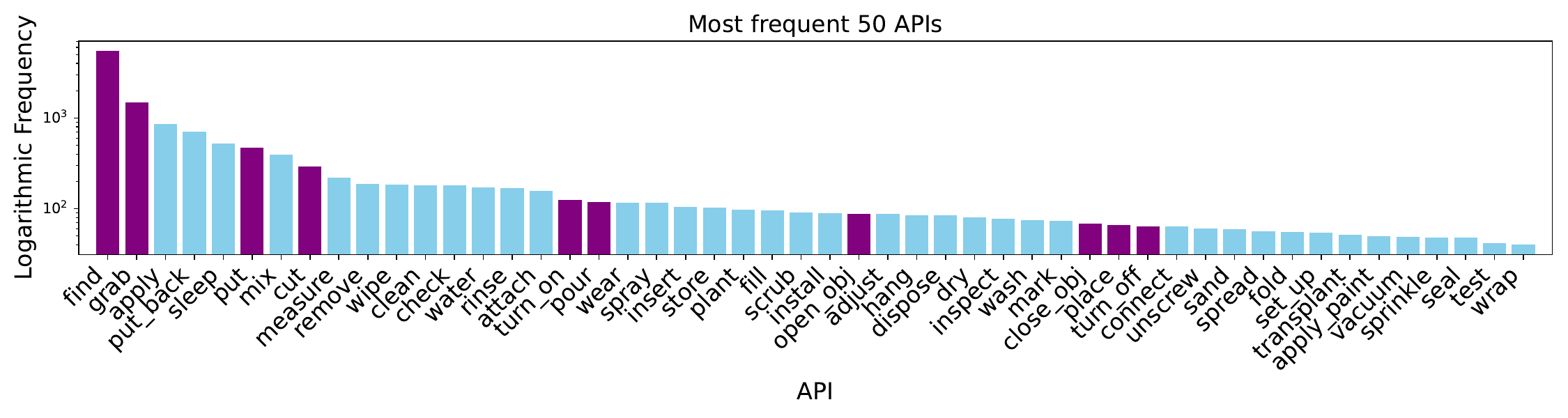}
    \caption{Top-50 most frequent APIs in the induced action space, with frequency in log scale. We use \textcolor{purple}{\rule{10pt}{5pt}}} to mark the APIs with exact/overlapping affordance to the primitive actions in existing embodied environments \citep{alfred, Puig2018VirtualHomeSH} and use \textcolor{infra-white}{\rule{10pt}{5pt}} to mark APIs that are beyond the action space of exiting environments.
    \label{fig:top_freq_apis}
\end{figure*}

\subsection{Evaluation Metrics}
Since our main contributions are based on the thought experiments, we focus on approximating the action space grounded in the hypothetical world (\S\ref{sec:worldapi}), instead of implementing a simulation of it. An exact execution-based evaluation is beyond the scope of our work. We thus approximate the execution-based evaluation with human evaluations and characterize the resulting action space through the lens of several automatic statistics.

\paragraph{Human Evaluation.} Since we do not have a simulator to execute the induced APIs and policies, we manually evaluate the quality of programs and APIs for the first 50 tutorials produced by each pipeline variant. We define two metrics to measure the quality of APIs and programs:

\underline{Redundancy:} In principle, we want the resulting pool of APIs to have low redundancy: each API should be atomic and unique in its functionality. 
For each new API, we quantify its redundancy with a $0-0.5-1$ scale measurement. An API is considered \emph{fully redundant} with a score of 1 if there exist straightforward composition(s) of existing APIs that can replicate the functionality of that new API; \emph{fully useful} with a score of 0 if there does not exist such compositions; and \emph{partially redundant/useful} with a score of 0.5 for all the other cases - which includes complicated action where decomposition is complicated without expert knowledge (e.g. install electrical conduit), synonyms that are similar to existing APIs semantically but require different low-level physical specifications (e.g. wipe vs. scrub, cut vs. snip ).

\underline{Faithfulness:} We approximate the simulator execution-based evaluation of generated programs with faithfulness measurement of $0-0.5-1$ scale. We define faithfulness on each individual instruction step: the objects are assumed to be in good initial conditions (e.g. the goal condition of successfully performing the previous step) at the beginning of each instruction step. Under this setup, a subprogram of its corresponding instruction step is \emph{fully faithful} with a score of 1  if all the goal conditions will be met upon execution; 0.5 if only part of the goal conditions can be met, and 0 if non of the goal condition will be met. In addition to the absolute scores, we also evaluate the faithfulness across pipeline variants based on relative ranking. For each instruction, we ask the annotators to rank the code snippets of different configurations based on their relative faithfulness.

\paragraph{Automatic evaluation.} \label{paragraph: auto_eval} We further quantify the characteristics of the induced API pool with the following statistics:\\
\begin{itemize}[leftmargin=10pt]
    \item 
\textbf{Size of API Pool} that records the total number of unique APIs defined and induced after each induction step.
 \item  \noindent\textbf{New API Induction Ratio} that measures the fraction of induced APIs over all the unique APIs evoked at each induction step. 
 \item  \noindent\textbf{API frequency} that counts the number of calls on each API over the full induction
\end{itemize}

\section{Results \& Analysis}
\subsection{Human Evaluation}\label{subsec:human_eval}
Table \ref{tab:qual_new} shows the human evaluation results of running the induction pipeline variants on the first 50 wikiHow tutorials from the 1000 test samples. In addition to the absolute redundancy score, we also report the redundancy scores when converting all the \texttt{partially redundant} APIs that are either too complicated to be further decomposed without expert knowledge (-Complex), or synonyms to existing APIs (-Synonym) to \texttt{fully useful}. As we believe in practice the further optimization of these APIs should be done at the implementation level. As indicated by the results, adding API use cases as demonstrations significantly reduces the number of new APIs induced and also decreases the redundancy. Further incorporating detailed wikiHow descriptions improves the faithfulness of the generated program.

\subsection{Automatic Evaluation}
\label{subsec:auto_eval}
We illustrate in Figure \ref{fig:freq_based} the expansion of the API pool while iteratively inducing on the sampled 1000 wikiHow tutorials, with filtering of infrequent APIs. 
At all levels of filtering, the pool of APIs demonstrates a diminishing increase in size and reaches a plateau after $\sim$ 600 steps of inductions. This indicates the induction pipeline approximates the \textbf{stabilized} space of frequently-evoked primitive actions to be $\sim 200 - 400$. Moreover, the saturation of the action space also indicates that as induction proceeds, LLMs shift from frequently inducing new APIs to largely reusing existing APIs. 
In Figure \ref{fig:induction_rate}, we provide the average ratio of induced APIs at each step of the induction.

Figure \ref{fig:top_freq_apis} displays the top 50 most frequently evoked APIs in the induced action space. It can be observed that the induced APIs cover a wide variety of physical actions that are necessary for carrying out the rich space of tasks in the real world. It is worth noting that the induction is only performed on a diverse yet very small fraction ($< 0.5\%$) of wikiHow tutorials. And with such limited scale yields a space of hundreds of APIs. The find suggests that the approximated $\sim 200 - 400$ frequent primitive actions can only serve as a proxy of the lower bound of a potentially much larger action space reflected in wikiHow. See Figure \ref{fig:visualization_apis} for t-SNE-based visualization of APIs based on BERT embeddings. 

Furthermore, we highlight with \textcolor{purple}{\rule{10pt}{5pt}} the induced APIs whose affordances are covered/supported by primitive actions of existing embodied environments \citep{alfred, Puig2018VirtualHomeSH}. The limited coverage (9 out of 50) provides evidence of the gap between action spaces of simulations and the real world, and motivates future work on simulations with richer primitive actions.

\section{Conclusion and Limitations}
Here, we attempt to approximate the size and properties of the action space of a versatile embodied agent that can carry out diverse tasks in the real world. We design a thought experiment with an induction pipeline to jointly induce primitive actions as APIs and agent policies as Pythonic programs, based on the hypothetical environments defined with wikiHow tutorials. Human and automatic evaluation verifies the usability of our induced APIs and provides an approximation of the lower bound of the realistic action space (300+). Moreover, our analysis highlights the deficiency in action diversity of existing embodied environments and motivates the development of action-rich simulations.

On the other hand, the induced action space is still noisy with a relatively high portion of redundant APIs, and it is challenging to scale up the evaluation without realizing APIs and executing policies induced by our pipeline. These limitations motivate future work in several interesting directions including reducing redundancy via self-correction \citep{madaan2023self, Jiang2024SELFINCORRECTLS} and evaluation in video-based embodied simulations \citep{Du2023LearningUP}.
\section*{Acknowledgements}
This work is in-part supported by ONR grant N00014-241-2089, and generous gifts
from Amazon and the Allen Institute for AI.
We are also grateful to the broader JHU CLSP community and our anonymous reviewers  for their support and constructive feedback 

\newpage
\onecolumn
\section{Reproducibility Checklist}
\appendix
This paper:
\begin{itemize}
\item Includes a conceptual outline and/or pseudocode description of AI methods introduced (yes)

\item Clearly delineates statements that are opinions, hypothesis, and speculation from objective facts and results (yes)

\item Provides well marked pedagogical references for less-familiare readers to gain background necessary to replicate the paper (yes)

\end{itemize}
Does this paper make theoretical contributions? (no)

\noindent Does this paper rely on one or more datasets? (yes)

\noindent If yes, please complete the list below.

\begin{itemize}
\item A motivation is given for why the experiments are conducted on the selected datasets (yes)

\item All novel datasets introduced in this paper are included in a data appendix. (NA)

\item All novel datasets introduced in this paper will be made publicly available upon publication of the paper with a license that allows free usage for research purposes. (NA)

\item All datasets drawn from the existing literature (potentially including authors’ own previously published work) are accompanied by appropriate citations. (yes)

\item All datasets drawn from the existing literature (potentially including authors’ own previously published work) are publicly available. (yes)

\item All datasets that are not publicly available are described in detail, with explanation why publicly available alternatives are not scientifically satisficing. (NA)
\end{itemize}

\noindent Does this paper include computational experiments? (yes)

If yes, please complete the list below.
\begin{itemize}
\item Any code required for pre-processing data is included in the appendix. (no).

\item All source code required for conducting and analyzing the experiments is included in a code appendix. (no)

\item All source code required for conducting and analyzing the experiments will be made publicly available upon publication of the paper with a license that allows free usage for research purposes. (yes)

\item All source code implementing new methods have comments detailing the implementation, with references to the paper where each step comes from (yes)

\item If an algorithm depends on randomness, then the method used for setting seeds is described in a way sufficient to allow replication of results. (no)

\item This paper specifies the computing infrastructure used for running experiments (hardware and software), including GPU/CPU models; amount of memory; operating system; names and versions of relevant software libraries and frameworks. (no)

\item This paper formally describes evaluation metrics used and explains the motivation for choosing these metrics. (yes/partial/no)

\item This paper states the number of algorithm runs used to compute each reported result. (no)

\item Analysis of experiments goes beyond single-dimensional summaries of performance (e.g., average; median) to include measures of variation, confidence, or other distributional information. (yes)

\item The significance of any improvement or decrease in performance is judged using appropriate statistical tests (e.g., Wilcoxon signed-rank). (no)

\item This paper lists all final (hyper-)parameters used for each model/algorithm in the paper’s experiments. (yes)

\item This paper states the number and range of values tried per (hyper-) parameter during development of the paper, along with the criterion used for selecting the final parameter setting. (no)
\end{itemize}
\newpage

\section{Annotation Guideline}
\subsection{Defining Objects} \label{subsec_appx: define objects}
Below is the Pythonic object definition of the \texttt{WorldObject} class.
\begin{lstlisting}[language=Python, caption=Object Definition, style=myCustomPythonStyle, numbers=none]
# Pythonic object definition
class WorldObject:
    def __init__(self,
                 obj_id: str,
                 actionable_properties: List[str],
                 material_properties: Dict[str, Any],
                 receptacles: List[Any],
                 receptacles_of: List[Any]
                 ):
        self.id = obj_id
        # actionable properties: properties that define what API calls can take this object as an argument
        self.actionable_properties = actionable_properties
        # material properties: properties that describe the physical attributes of the object
        # can be manipulated by api calls
        self.material_properties = material_properties
        # the containers of this object
        self.receptacles = receptacles
        # other objects that are contained in the current object
        self.receptacles_of = receptacles_of
\end{lstlisting}
\subsection{Defining Primitive actions (APIs)}
\subsubsection{Base APIs} \label{subsubsec_appx: base apis}
Here is the list of base APIs we defined, similar to existing simulators like \cite{Puig2018VirtualHomeSH, padmakumar2021teach}.
\begin{lstlisting}[language=Python, caption=base APIs, style=myCustomPythonStyle, numbers=none]
# Navigation: find
# Navigate to the vicinity (interactable distance) of the target object (e.g. chocolate bar)
find(obj=chocolate_0)

# Picking up and putting down objects: grab, put
# pick up objects
grab(obj=cup_0)
# put down grabbed objects on/into the target object
put(obj=cup_0, target=sink_0)

# Basic interactions with objects: open_obj, close_obj, turn_on, turn_off, 
# open and close the target object
open_obj(obj=microwave_owen_0)
close_obj(obj=microwave_owen_0)

# switch on and off the target object, with optional arguments such as power level
turn_on(obj=burner_0, power="high")
turn_off(obj=burner_0)

# API that turns the agent into standby for a given duration, e.g. waiting for the current process: sleep
sleep(duration="300s")

# Putting the target object back to its original receptacles
put_back(obj=plate_0)
\end{lstlisting}

\subsubsection{Granularity} \label{subsubsec_appx: granularity}
\paragraph{Granularity - Avoiding overly-abstract actions:} Many of the wikiHow instructions describe scenarios/situations in which the agent is expected to participate and perform certain action(s). For example ``Dry the denim using high heat.'' One straightforward way of creating APIs would be a parsing-like method that identifies the predicate(s) in these situations and converts them into API definitions accordingly:
\begin{lstlisting}[language=Python, caption=Situation-based API definition, numbers=none]
# Dry the denim using high heat.
dry(obj=denim_0, dryer=dryer_0, temperature="high")
\end{lstlisting}
However, we argue that it is not a desired approach to create the action space since it degrades the agent planning task into semantic parsing (e.g. the \texttt{dry} API can be viewed as the \texttt{Cause\_to\_be\_dry} frame in FrameNet (cite here), with \texttt{dryer\_0} be the \texttt{Instrument} frame element). While linguistically grounded, it creates APIs at a problematic level of granularity from an action perspective - drying something with a dryer requires a totally different set of actions to drying something with iron or air-drying. In practice, it causes an extra layer of confusion and complexity in implementing/learning the API for a simulated/real robot.

We instead adopt an \textit{affordance-based} approach: whenever it is possible, define APIs according to how can the agent interact with the objects based on their affordance (e.g. the dryer can be opened and turned on to perform drying)
\begin{lstlisting}[language=Python, caption=Affordance-based API definition, numbers=none]
# Dry the denim using high heat.
open_obj(obj=dryer_0)
put(obj=denim_0, target=dryer_0)
close_obj(obj=dryer_0)
turn_on(obj=dryer_0, power="high", duration="1200s")
sleep(duration='1200s')
\end{lstlisting}

\paragraph{Granularity - Avoiding too low-level actions:} We aim at defining the APIs to maximize atomicity in a decomposition-based manner. Whenever it is possible, we decompose a candidate API into sub-actions that are more atomic. To an extreme, this always leads to decomposing any API into low-level controls. For example, ``pour the water into the bowl'' and ``wipe the table'' can be annotated as AI2-ThOR type of controls:
\begin{lstlisting}[language=Python, caption=Decomposition into low-Level controls, style=myCustomPythonStyle, numbers=none,]
# pour the water into the bowl.
grab(cup_0)
find(bowl_0)
move_held_object(ahead=0.0, right=0.0, up=0.5)
rotate_held_object(pitch=0, yaw=0, roll=45)

# wipe the table
grab(paper_towel_0)
find(table_0)
move_held_object(ahead=0.0, right=0.0, up=-0.5)
move_held_object(ahead=0.5, right=0.5, up=0.0)
move_held_object(ahead=-1.0, right=-1.0, up=0.0)
\end{lstlisting}
This degrades the action space into just a handful of control APIs with complicated spatial-temporal argument realization, which is infeasible in our hypothetical environment without explicit spatial relation specification, and harms the reusability of learned APIs in practice. We thus regulate the decomposition process such that any candidate API is considered as \texttt{atomic} (thus eligible to be added to the action space) if any further decomposition involves low-level controls. Accordingly, for the examples above we define the following APIs
\begin{lstlisting}[language=Python, caption=Decomposition at proper level of Granularity]
# pour the water into the bowl.
grab(cup_0)
find(bowl_0)
pour(obj=cup_0, target=bowl_0)

# wipe the table
grab(paper_towel_0)
find(table_0)
wipe(obj=table_0, tool=paper_towel_0)
\end{lstlisting}

\subsubsection{State Checking and Feedback Loop} \label{subsubsec_appx: feedback}
In addition to linear chains of actions, agent planning also involves collecting feedback from the environment and performing actions accordingly. Similar to prior work, we adapt feedback loops to check the object states and carry out conditional actions. 
\begin{lstlisting}[language=Python, caption=Feedback loop, style=myCustomPythonStyle, numbers=none]
# if the chocolate is not smooth, heat it in the microwave for 10 seconds
if chocolate_0.material_properties['form'] != 'smooth':
    find(obj=microwave_owen_0)
    open_obj(obj=microwave_owen_0)
    put(obj=bowl_0, target=microwave_owen_0)
    close_obj(obj=microwave_owen_0)
    turn_on(obj=microwave_owen_0, power="low", duration="10s")
\end{lstlisting}

\begin{figure*}[ht]
    \centering
    \includegraphics[scale=0.72,trim=0cm 2.0cm 4.5cm 1.0cm,clip,width=\textwidth]{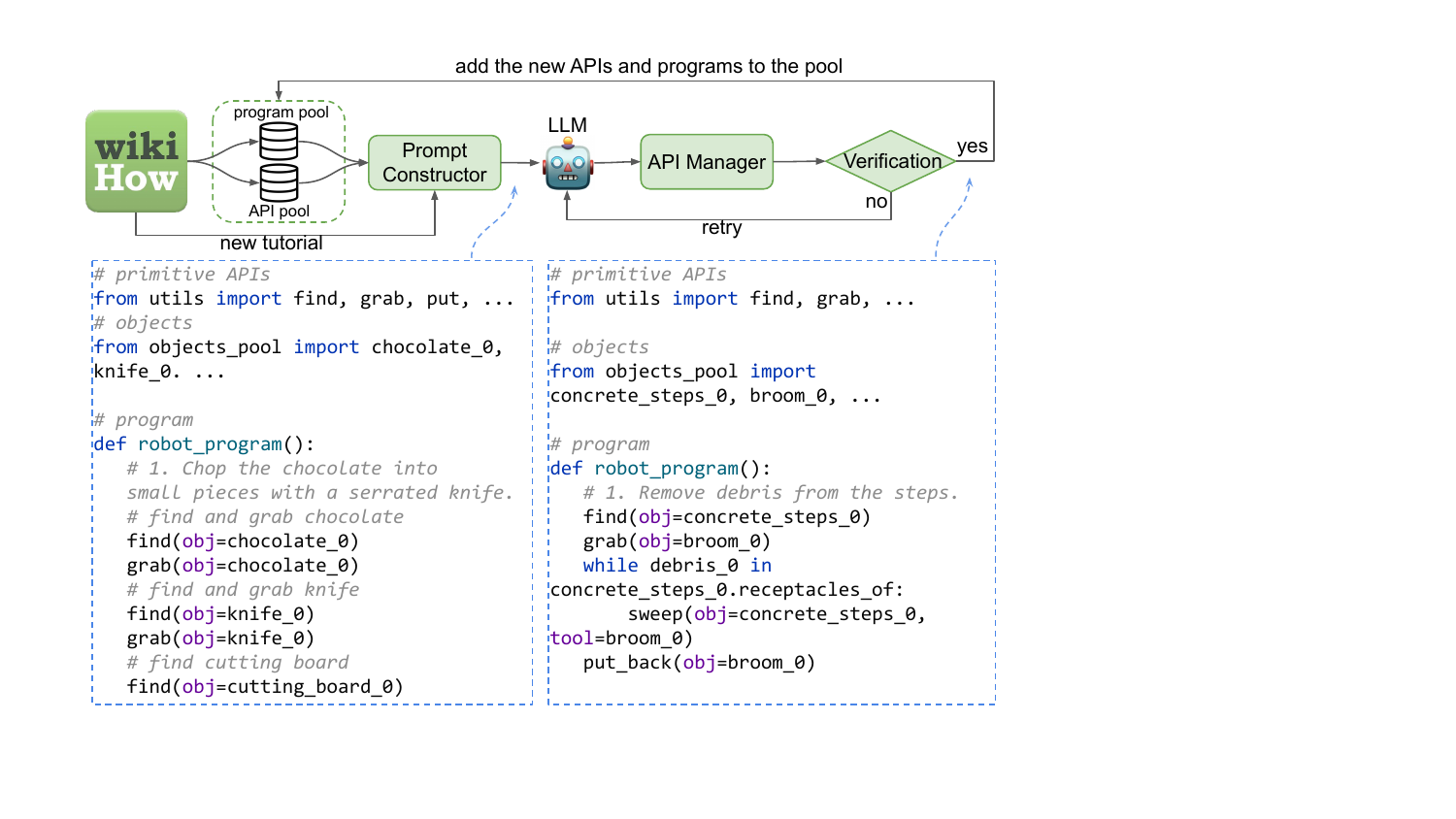}
    \caption{Proposed pipeline that jointly induces new APIs and programs.}
    \label{fig:pipeline:appendix}
\end{figure*}

\begin{figure*}[ht]
    \centering
    \includegraphics[scale=0.35,width=0.5\textwidth]{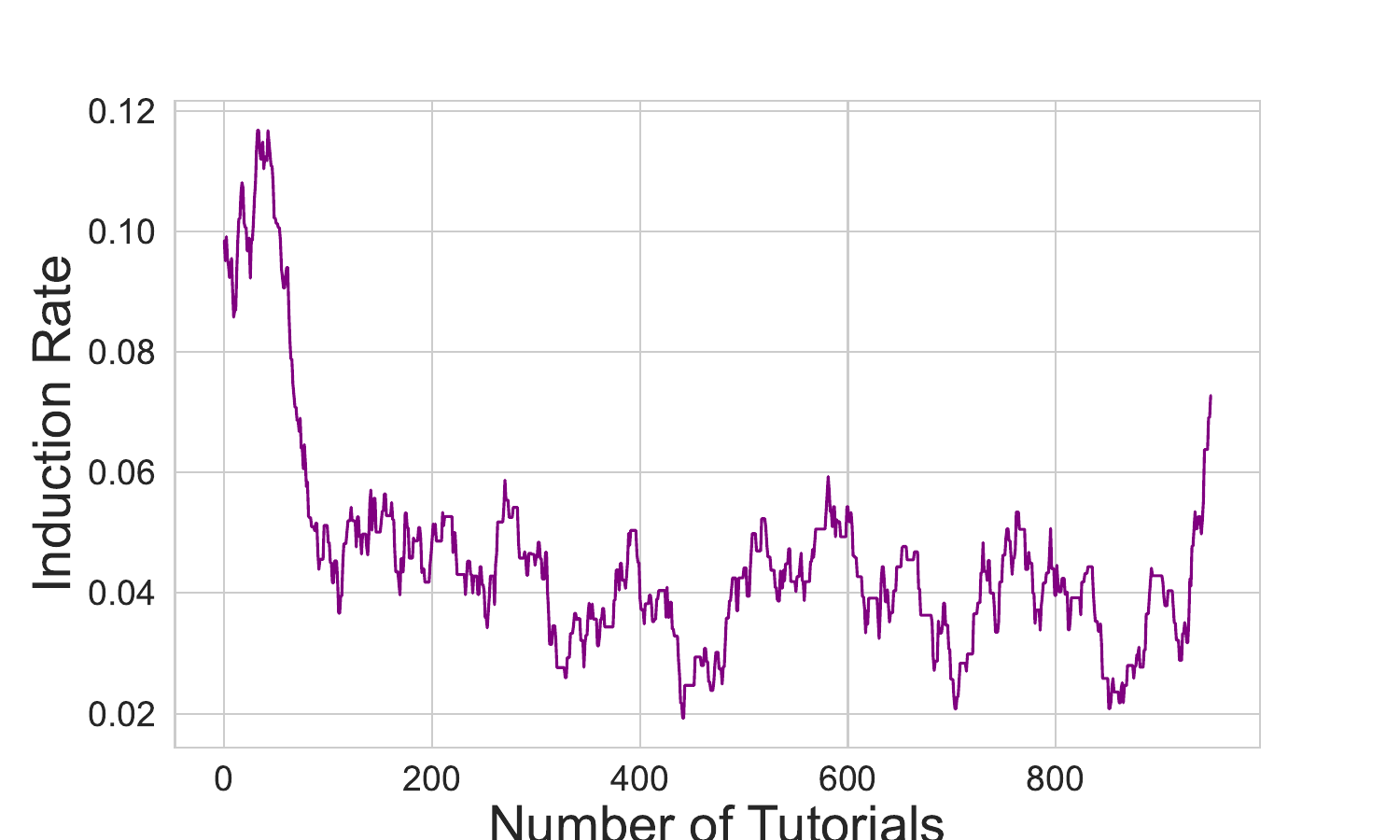}
    \caption{Moving average induction ratio with a window size of 50 tutorials. For an induction step, the induction ratio is defined as the fraction of hallucinated new APIs over the total number of unique APIs being evoked in the generated program. We calculate the moving average induction rate with a look-ahead window with a size of 50 steps. The spike at the tail is caused by the regeneration with the tutorials with a high initial induction ratio.}
    \label{fig:induction_rate}
\end{figure*}

\begin{figure*}[ht]
    \centering
    \includegraphics[trim=0cm 0cm 0cm 0cm,width=0.9\textwidth]{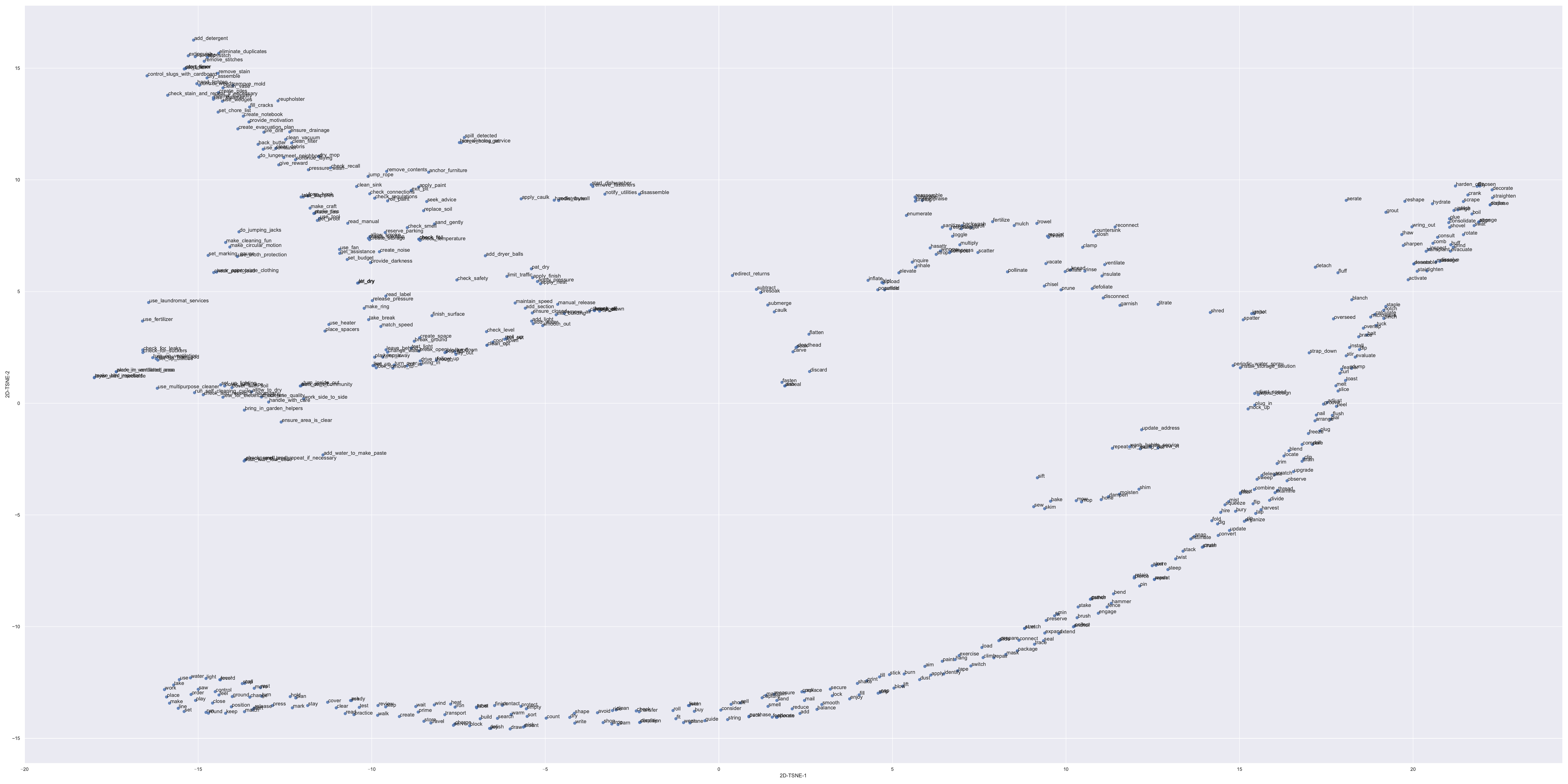}
    \caption{t-SNE representation of the APIs via BERT embeddings.}
    \label{fig:visualization_apis}
\end{figure*}

\end{document}